\documentclass[letterpaper, 10 pt, conference]{ieeeconf}  

\IEEEoverridecommandlockouts                              

\usepackage{graphicx} 
\usepackage{times} 
\usepackage{amsmath} 
\usepackage{amssymb}  

\usepackage{float}
\usepackage{bm}
\usepackage{cite}

\usepackage{color}
\usepackage{subfigure}

\usepackage{hyperref}
\hypersetup{
	colorlinks=true,
	linkcolor=blue,
	filecolor=blue,      
	urlcolor=blue,
	citecolor=cyan,
}

\usepackage[linesnumbered,ruled]{algorithm2e}

\title{\LARGE \bf
Boundary Effect-Aware Visual Tracking for UAV with Online Enhanced Background Learning and Multi-Frame Consensus Verification
}

\author{Changhong Fu$^{1,*}$, Ziyuan Huang$^{2}$, Yiming Li$^{1}$, Ran Duan$^{3}$, and Peng Lu$^{3}$ 
\thanks{$^{1}$Changhong Fu and Yiming Li are with the School of Mechanical Engineering, Tongji University, 201804 Shanghai, China
        {\tt\small changhongfu@tongji.edu.cn}}%
\thanks{$^{2}$Ziyuan Huang is with the School of Automotive Studies, Tongji University, 201804 Shanghai, China
		{\tt\small tjhuangziyuan@gmail.com}}%
\thanks{$^{3}$Ran Duan and Peng Lu are with the Adaptive Robotic Controls Lab (ArcLab), Hong Kong Polytechnic University (PolyU), Hong Kong, China
       {\tt\small peng.lu@polyu.edu.hk}}%
}
\begin{document}
\maketitle
\thispagestyle{empty}
\pagestyle{empty}
\begin{abstract}
Due to implicitly introduced periodic shifting of limited searching area, visual object tracking using correlation filters often has to confront undesired boundary effect. As boundary effect severely degrade the quality of object model, it has made it a challenging task for unmanned aerial vehicles (UAV) to perform robust and accurate object following. Traditional hand-crafted features are also not precise and robust enough to describe the object in the viewing point of UAV. In this work, a novel tracker with online enhanced background learning is specifically proposed to tackle boundary effects. Real background samples are densely extracted to learn as well as update correlation filters. Spatial penalization is introduced to offset the noise introduced by exceedingly more background information so that a more accurate appearance model can be established. Meanwhile, convolutional features are extracted to provide a more comprehensive representation of the object. In order to mitigate changes of objects' appearances, multi-frame technique is applied to learn an ideal response map and verify the generated one in each frame. Exhaustive experiments were conducted on 100 challenging UAV image sequences and the proposed tracker has achieved state-of-the-art performance.
\end{abstract}
\section{INTRODUCTION}
\label{sec:INTRODUCTION}
Visual object tracking plays a crucial role in unmanned aerial vehicle (UAV) applications such as obstacle avoidance \cite{fu2014ICRA}, wild-life monitoring \cite{olivares2015Sensors} and object following \cite{mueller2016IROS} (e.g. humans, cars, boats, etc.). Generally, UAV object tracking demands the tracker to keep track of an visual object detected by UAV.
Due to online nature of learning as well as appearance changes of object, such as deformation, occlusion and illumination variation, object tracking remains a challenging task despite of recent significant progress. Especially in UAV scenarios, frequently occurrence of viewpoint change, fast movement and camera motion adds to its difficulty \cite{mueller2016ECCV}.
\begin{figure}[!ht]
	\centering  
	\includegraphics[scale=0.33]{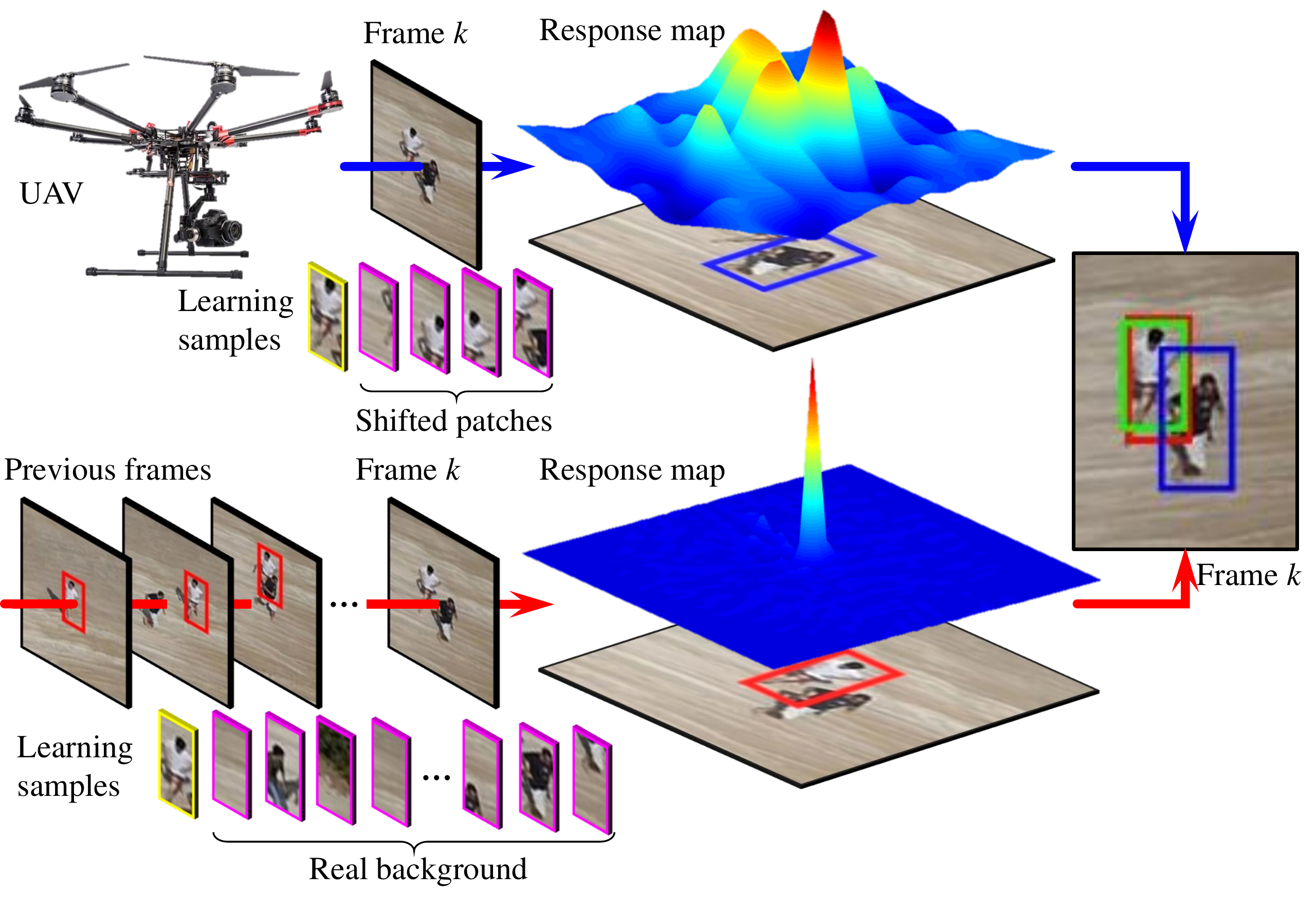}
	\caption{Comparison between response maps of other CF based trackers and proposed tracker. Yellow samples are served as positive samples and pink ones are negative samples. Green bounding box in the right image is the ground truth. Red and blue boxes are from respectively the proposed tracker and other trackers. Real background are introduced in the proposed tracker as negative samples. Enhanced background learning and response map verification have notably suppressed the noise response in the background.}
	\label{img:introduction}
\end{figure}

Correlation filter (CF) based framework has been successfully applied to tackle the aforementioned difficulties in recent years \cite{kiani2017ICCV, danelljan2015ICCV, henriques2015TPAMI}, due to its computational efficiency and sufficient tracking performance. It learns a correlation filter from the training samples online and calculates the correlation response in the frequency domain. The filter is then applied in the new frame to generate a response map. The object is located in this frame where the response value is the highest.

Learning CF efficiently requires, however, circular shifting operation on the training and detection samples. This generates periodic representation extension of these samples, which leads to undesired boundary effects since the search region is limited, see Fig. \ref{img:introduction}. Several approaches have been taken to mitigate boundary effects \cite{kiani2017ICCV, danelljan2015ICCV}. Further, with limited training samples, learned models might be over-fitted and lack of negative patches will damage the accuracy of the tracker as well. In addition, many trackers based on CFs use hand-crafted features. Histogram of oriented gradients (HOG) is one of the most popular among those features \cite{henriques2015TPAMI, kiani2017ICCV, danelljan2015ICCV}. Recently, convolutional-features-based trackers have demonstrated their competence in more precisely describing the object and thus achieving state-of-the-art results in visual tracking \cite{qi2016CVPR, ma2015ICCV, wang2015ICCV}. 

The information extracted from the response map has not received too much attention \cite{choi2017CVPR, wang2017CVPR}. However, it reveals to some extent the similarity between the model learned from previous consecutive frames and the newly detected target, which can be used to determine whether the object has differed drastically from the learned model and whether the detection result can be trusted.

This work focuses on solving boundary effect of CFs using enhanced background learning, achieving better representation of objects using convolutional features and estimate the consensus of tracking result depending on multi-frame consensus verification. A novel boundary effect-aware visual tracking approach is proposed, i.e. BEVT tracker.

Contributions of this work are listed as follows:
\begin{itemize}
	\item A novel enhanced background learning approach is presented to learn background information and suppress the noise introduced by background at the same time. BEVT densely extracts background samples to enlarge search window, and a penalization to the  object according to its spatial location is applied to suppress background noise. The effect can be seen in Fig. \ref{img:introduction}.
	\item A new feature extraction approach is applied. Different layers of CNN are exploited to provide both spatial and semantic information of objects, raising the preciseness of appearance models.
	\item A novel approach is utilized to raise the robustness of appearance model. 	
	A response map model learned from consecutive frames is compared with current-frame response map to generate a consensus score, which is used to influence the learning process so that unnecessary learning is avoided and necessary learning is enhanced.
	\item The tracker is evaluated on 100 challenging UAV image sequences and compared with other state-of-the-art trackers. Competitive accuracy is demonstrated in the experiments.
\end{itemize}

To the best of our knowledge, it is the first time that the presented BEVT tracker is designed in the literature and employed in UAV tracking.

\section{RELATED WORKS}
\label{sec:RELATEDWORKS}
\subsection{Tracking with correlation filter}
CF framework has been widely applied in the field of object tracking. Its success depends on the implicit inclusion of all periodically shifted learning sample and the exceeding computational efficiency. Many trackers use CF framework such as minimum output sum of squared error (MOSSE), kernelized correlation filters \cite{henriques2015TPAMI} and multiple other trackers \cite{van2009TIP, danelljan2014BMVC, kiani2017ICCV, danelljan2015ICCV}. However, correlation filter has a natural defect of boundary effect. Efforts have been made to investigate the problem. Spatially regularized DCF (SRDCF) utilizes spatial information to expand search regions \cite{danelljan2015ICCV}, and learning background-aware CF (BACF) use negative patches from background for the correlation filter to learn \cite{kiani2017ICCV}. However, with background introduced, more irrelevant information is also brought into the search window. 
\subsection{Tracking by convolutional feature}
Deep convolutional networks are receiving more attention in recent years. Some trackers directly exploit the idea of deep architecture and built upon it \cite{choi2017CVPR}, while others extract CNN feature within CF framework \cite{ma2015ICCV, danelljan2015ICCV}. Since object representation is crucial in object tracking of UAV due to its requirement of online training, the application of CNN features is rapidly expanding in this field due to its comprehensive description. Both single-layer \cite{danelljan2015ICCV} and multi-layer \cite{ma2015ICCV} features are investigated and applied in the CF framework. 
\subsection{Verification by response map}
Response map is the result generated by CF tracker in each frame and objects are believed to be where its value is the highest. Therefore, response map reveals certain information about the object to be tracked in this frame. Attentional correlation filter network for adaptive visual tracking (ACFN) uses response map to select suitable tracking module \cite{choi2017CVPR}. Large margin object tracking with circulant feature maps (LMCF) utilizes maximum response value and the average peak to correlation energy (APCE) to measure the confidence of each tracking result \cite{wang2017CVPR}.
\section{PROPOSED TRACKING APPROACH}
\label{sec:PROPOSEDTRACKINGAPPROACH}
\begin{figure*}[!ht]
	\centering
	\includegraphics[scale=0.45]{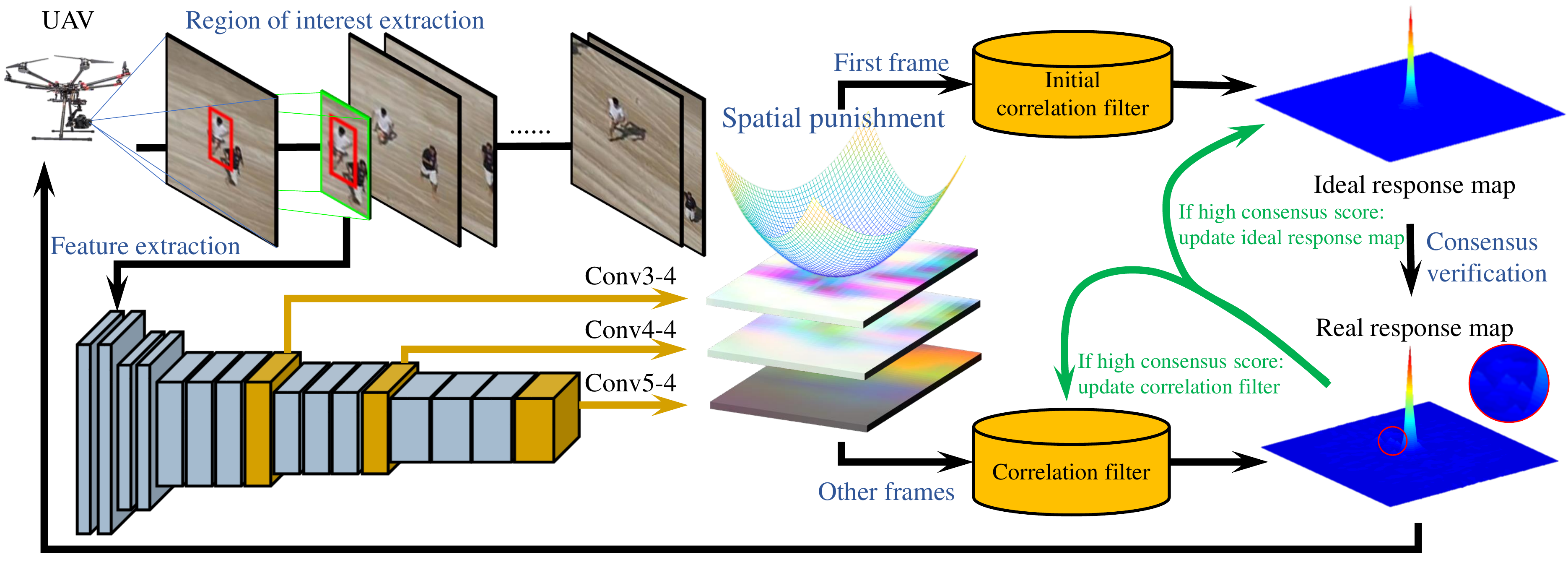}
	\caption{Main structure of the proposed BEVT tracker. Because of enhanced background learning, background noise is further suppressed. Therefore, details in the real response map is zoomed for better clarity.}
	\label{img:overview}
\end{figure*}

In this work, a novel correlation filter with enhanced background learning and multi-frame consensus verification is proposed. Its main structure is shown in Fig. \ref{img:overview}.
\subsection{Overall objective}
\label{subsec:OVERALLOBJECTIVE}
The overall objective of our tracker is to minimize the loss function with enhanced background learning, which can be expressed as the following equation:
	\begin{equation}\label{objective}
	\mathcal { E } ( \mathbf { w } ) = \frac { 1 } { 2 } \Vert \mathbf { y } - \sum _ { d = 1 } ^ { D } \mathbf { Bx }^d \star \mathbf { w } ^ { d } \Vert _2^2 + \sum _ { d = 1 } ^ { D } \Vert \mathbf { pw } ^ { d } \Vert _2^2 \ ,
	\end{equation}

\noindent where $\mathbf{y}\in\mathbb{R}^N$ denotes the desired correlation response, $\mathbf{x}^d\in\mathbb{R}^N(d=1,2,...,D)$ denotes the ${d}$th channel of vectorized input image and $\mathbf{w}^{d}\in\mathbb{R}^M$ denotes the ${d}$th channel of the correlation filter. $\mathbf{B} \in \mathbb{R}^{M\times N}$ is a binary matrix, adopted to crop the middle $M$ elements of signal $\mathbf{x}^d$. Moreover, to further mitigate the boundary effect, a penalization matrix $\mathbf{p}\in\mathbb{R}^M$ is introduced to emphasize the central part of the extracted feature.\\
\noindent\textbf{Remark 1}: The enhanced background learning not only can mitigate boundary effect by employing cropping matrix $\mathbf{B}$ to extract background samples, noise introduced by background is also suppressed in the proposed approach by spatial penalization $\mathbf{p}$. In addition, the feature vector $\mathbf{x}$ is extracted from VGG-Net \cite{simonyan2014arXiv} using conv3-4, conv4-4 and conv5-4 combined together, since early layers provide spatial details and deep layers contains semantic information to prevent deformations. Details are shown in the middle of Fig. \ref{img:overview}.

The aforementioned objective can be equally expressed as the following equation:
	\begin{equation}\label{matrix_form_of_objective}
		\mathcal { E } ( \mathbf { w } ) = \frac{1}{2} \sum_{n=1}^N \lVert \mathbf{y}(n) - 
		\sum_{d=1}^D \mathbf{w} ^{d\top} \mathbf{Bx}^{d}[\Delta\tau_n] \rVert _2^2 + 
		\sum_{d=1}^D \lVert \mathbf{pw}^{d} \rVert _2^2 \ ,
	\end{equation}

\noindent where ${\mathbf{y}(n)}$ is the ideal response and $\mathbf{Bx}^{d}[\Delta\tau_n]$ is the cropped input image sample. $[\Delta\tau_n]$ is the circular shift operator, and $\mathbf{Bx}^{d}[\Delta\tau_n]$ is the cropped input image sample that is shifted by n elements.The superscript $\top$ denotes the conjugate transpose of a complex vector or matrix.

All the summations can be expressed in matrix form, which gives, 
	\begin{equation}\label{simplification_of_objective}
		\mathcal { E } ( \mathbf { w } ) = \frac{1}{2}  \lVert \mathbf{y} - 
		\mathbf{X}  \left(\textbf{I}_D \otimes \mathbf{B}^{\top}\right) \mathbf{w}\rVert _2^2 + 
		\lVert \mathbf{\tilde{p}w} \rVert _2^2 \ , 
	\end{equation}
	
\noindent where 
$
\mathbf{y} = \left[ \begin{matrix} 
\mathbf{y}(1) \cdots \mathbf{y}(N) 
\end{matrix}\right] ^{\top}, $
$\mathbf{w} = \left[ \begin{matrix}
\mathbf{w}_1^{\top} \cdots \mathbf{w}_D^{\top}\
\end{matrix}\right]^{\top},$
$\mathbf{X} =  \left[ \begin{matrix} 
	\mathbf{x} [\Delta\tau_1]^{\top} \cdots  \mathbf{x} [\Delta\tau_N]^{\top} 
\end{matrix}\right]^{\top},$
and 
$
\mathbf{\tilde{p}} = \left[\begin{matrix}
\mathbf{p}^{\top}\ \mathbf{p}^{\top}\ \cdots \  \mathbf{p}^{\top}\
\end{matrix}
\right]^{\top}.
$
$\mathbf{I}_D \in \mathbb{R}^{D \times D}$ is an identity matrix and $\otimes$ is the operator of Kronecker product. \\
\noindent\textbf{Remark 2}: Simplification process can be seen in the Appendix.

\subsection{Transfer into frequency domain}
\label{subsec:TRANSFERINTOFREQUENCYDOMAIN}

To speed up the calculation, the correlation operations are normally carried out in the frequency domain. Therefore, (\ref{simplification_of_objective}) can be transferred into frequency domain to be the following equation:
	\begin{equation}\label{frequency_format}
	\begin{aligned}
		&\mathcal { \hat{E} } ( \mathbf { w }, \mathbf{\hat{g}} ) = \frac{1}{2}  \lVert \mathbf{\hat{y}} - 
		\mathbf{\hat{X}} \mathbf{\hat{g}}\rVert_2^2 + 
		\lVert \mathbf{\tilde{p}w} \rVert _2^2
		\\
		& s.t.\quad \hat{\mathbf{g}} = \sqrt{N}(\mathbf{I}_D \otimes \mathbf{F} \mathbf{B}^{\top} )\mathbf{w}
	\end{aligned} \ ,
	\end{equation}

\noindent where $\mathbf{ \hat{g} } \in \mathbb{C} ^{DN \times 1}$ is introduced in preparation for the further optimization operations. The superscript $ \hat{} $ indicates the discrete Fourier transform of a signal, i.e.,  $\hat{\bm{\alpha}} = \sqrt{N}{F}\bm{\alpha}$. Therefore,  $\mathbf{\hat{y}}$, $\hat{\mathbf{X}} = [ diag(\hat{\mathbf{x}}^1)^\top, \cdots, diag(\hat{\mathbf{x}}^D)^\top]$ and $\mathbf{\hat{g}}$ is respectively the Fourier form of $\mathbf{y}$, $\mathbf{X}$ and $\mathbf{g} = \left(\textbf{I}_D \otimes \mathbf{B}^{\top}\right) \mathbf{w}$ in (\ref{simplification_of_objective}).

\subsection{Optimization operations}
\label{subsec:OPTIMIZATIONOPERATIONS}
To fully exploit the convexity of (\ref{frequency_format}), we take advantage of Alternative direction method of multipliers (ADMM)\cite{boyd2011FTML} to calculate the optimal solution to the equation. First, we transform (\ref{frequency_format}) to Augmented Lagrangian format as the following equation:
	\begin{equation} \label{Lagragian_format}
		\begin{aligned}
			\mathcal{L} (\mathbf{w}, \hat{\mathbf{g}}, \hat{\zeta}) = 
			& \frac{1}{2}  \lVert \hat{\mathbf{y}} - \hat {\mathbf{X}} \hat{\mathbf{g}}\rVert _2^2 + 
			\lVert \mathbf{\tilde{p}w} \rVert _2^2 
			\\
			& + \hat{\zeta}^{\top}(\hat{\mathbf{g}} - \sqrt{N} \left(\mathbf{I}_D \otimes \mathbf{FB}^{\top}\right) \mathbf{w}) 
			\\
			& + \frac{\mu}{2} \lVert \hat{\mathbf{g}} - \sqrt{N} \left(\mathbf{I}_D \otimes \mathbf{FB}^{\top}\right) \mathbf{w} \rVert^2
		\end{aligned} \ ,
	\end{equation}

\noindent where the penalty factor is $\mu$ and the Lagrangian vector in the Fourier domain $\hat{\bm\zeta} \in \mathbb{C}^{DN\times 1}$ is defined as $\hat{\bm\zeta} = [\hat{\bm\zeta}^{1\top}, \cdots, \hat{\bm\zeta}^{D\top}]^\top$. By employing ADMM, the equation should be divided into two subproblems, respectively the following $\mathbf{\hat{g}^*}$ and $\mathbf{w}^*$. Then (\ref{Lagragian_format}) can be iteratively solved by solving these two subproblems alternatively. 

\subsubsection{Subproblem $\mathbf{w}^*$}
\label{subsubsec:SUBPROBLEMW*}
	\begin{equation}\label{subproblem_w}
		\begin{aligned}
		\mathbf{w}^* &=\arg \min _ { \mathbf { w } } \Big\lbrace
		\lVert \mathbf{\tilde{p}w} \rVert ^2 + \hat{\zeta}^{\top}(\hat{\mathbf{g}} - \sqrt{N} \left(\mathbf{I}_D \otimes \mathbf{FB}^{\top}\right) 
		\\
		& \quad + \frac{\mu}{2} \lVert \hat{\mathbf{g}} - \sqrt{N} \left(\mathbf{I}_D \otimes \mathbf{FB}^{\top}\right) \mathbf{w} \rVert _2^2
		\Big\rbrace
		\\
		&= \left( \frac{2\mathbf{\tilde{p}}^{\top} \mathbf{\tilde{p}}}{N} + \mu \right)^{-1}
		\left(\zeta + \mu \mathbf{g} \right)
		\end{aligned} \ ,
	\end{equation}

\noindent where $\mathbf{g}$ and $\zeta$ can be calculated from $\mathbf{\hat{g}}$ and $\hat{\zeta}$ respectively through the following Inverse Fast Fourier Transform (IFFT) equations:
	\begin{equation}
	\left\{
	\begin{aligned}
	\mathbf{g} & = & \frac { 1 } { \sqrt { N } } \left(\mathbf { I } _ { D } \otimes \mathbf { FB } ^ { \top } \right) \hat { \mathbf { g } } \\
	\mathbf { \zeta } & = & \frac { 1 } { \sqrt { N } } \left(  \mathbf { I } _ { D } \otimes \mathbf { FB } ^ { \top } \right) \hat { \mathbf { \zeta } } \\ 
	\end{aligned}
	\right. \ .
	\end{equation}

\noindent\textbf{Remark 3}: Note that inside the first pair of brackets in the calculation result of (\ref{subproblem_w}), $\mathbf{\tilde{p}}^{\top} \mathbf{\tilde{p}}$ and $N$ are constant, and parameter $\mu$ is preset for each iteration before tracking, so the first item of $\mathbf{w} ^*$ can be pre-calculated before running the algorithm so that the inverse operation won't affect the efficiency.

\subsubsection{Subproblem $\mathbf{g}^*$}
\label{subsubsec:SUBPROBLEMG*}
	\begin{equation} \label{subproblem_g}
	\begin{aligned}
	\hat{\mathbf{g}}^* = &\arg \min_{ \mathbf { g } }  \Big\lbrace
	\frac{1}{2}  \lVert \hat{\mathbf{y}} - 
	\hat {\mathbf{X}} \hat{\mathbf{g}}\rVert^2 
	\\
	&+ \hat{\zeta}^{\top} \left(\hat{\mathbf{g}} - \sqrt{N} \left(\mathbf{I}_D \otimes \mathbf{FB}^{\top}\right) \right)
	\\
	&+\frac{\mu}{2} \lVert \hat{\mathbf{g}} - \sqrt{N} \left(\mathbf{I}_D \otimes \mathbf{FB}^{\top}\right) \mathbf{w} \rVert _2^2
	\Big\rbrace
	\end{aligned} \ .
	\end{equation}

Calculating (\ref{subproblem_g}) directly would be time-consuming but fortunately, $\mathbf{\hat{X}}$ is sparse banded, which means the computation can be accelerated through separately calculating each element of $\mathbf{\hat{y}}$, i.e. $\mathbf{\hat{y}}(n), n=1,2,...,N$ because each $\mathbf{\hat{y}}$ relies solely on $\hat{\mathbf{x}}(n)=\left[\hat {\mathbf{x}}^1(n),\hat {\mathbf{x}}^2(n),...,\hat {\mathbf{x}}^D(n)\right]^{\top}$
and $\hat { \mathbf { g } } ( n ) = \left[ \operatorname { conj } \left( \hat { \mathbf { g } } ^ { 1 } ( n ) \right) , \ldots , \operatorname { conj } \left( \hat { \mathbf { g } } ^ { D } ( n ) \right) \right] ^ { \top }$, where conj(.) indicates the complex conjugate operation. Hence, subproblem $\mathbf{g}^*$ can be divided as $N$ independent sub-subproblems, which are to be solved over $n=[1,2,...N]$:
	\begin{equation}
		\begin{aligned}
		\hat{\mathbf{g}}(n)^* = &\arg \min_{ \mathbf { g } }  \Big\lbrace
		\frac{1}{2}  \lVert \hat{\mathbf{y}}(n) - \hat {\mathbf{x}}(n)^{\top} \hat{\mathbf{g}}(n)\rVert _2^2 
		\\
		&+ \hat{\zeta}(n)^{\top}\left( \hat{\mathbf{g}}(n) - \hat{\mathbf{w}}(n) \right)
		\\
		&+\frac{\mu}{2} \lVert \hat{\mathbf{g}}(n) -  \hat{\mathbf{w}}(n) \rVert^2
		\Big\rbrace
		\end{aligned} \ ,
	\end{equation}
\noindent where $\hat { \mathbf { w } } ( n ) = \left[ \hat { \mathbf { w } } ^ { 1 } ( n ) , \ldots , \hat { \mathbf { w } } ^ { D } ( n ) \right]$ and $\hat { \mathbf { w } } ^ { d } = \sqrt { D } \mathbf { F } \mathbf { P } ^ { \top } \mathbf { w } ^ { d }$. The solution to each sub-subproblem can be expressed as:
	\begin{equation}\label{solution_to_sub-subproblem}
	\begin{aligned}
	\hat{\mathbf{g}}(n)^* = &\left( \hat{\mathbf{x}}^{\top}(n) \hat{\mathbf{x}}(n) + \mu \mathbf{I}_D\right)^{-1}
	\\
	&\left( \hat{\mathbf{x}}^{\top}(n) \hat{\mathbf{y}}(n) - \hat{\zeta}(n) + \mu \hat{\mathbf{w}}(n) \right)
	\end{aligned} \ .
	\end{equation}

Unfortunately, the inverse operation here will take a huge amount of time to be carried out. In order to improve the efficiency, the Sherman-Morrison formula \cite{sherman1950AMS} is applied, which is $(\mathbf{A} + \mathbf{u} \mathbf{v}^{\top}) ^ {-1} = \mathbf{A} ^ {-1} - \mathbf{A}^{-1} \mathbf{u} (\mathbf{I}_k + \mathbf{v}^{\top} \mathbf{A}^{-1} \mathbf{u}) ^ {-1} \mathbf{v}^{\top} \mathbf{A}^{-1} $ in generalization form, where matrix $u$ is an $n \times k$ matrix, $v$ is an $k \times n$ matrix and A is an $n \times n$ matrix. In our case, $A = \mu \mathbf{I}_D$ and $\mathbf{u} = \mathbf{v} = \mathbf{\hat{x}}(n)$. Therefore, (\ref{solution_to_sub-subproblem}) can be equally expressed as:
	\begin{equation}\label{subproblem_g_result}
	\begin{aligned} 
	\hat { \mathbf { g } } ( n ) ^ { * } = & \frac { 1 } { \mu } ( \hat { \mathbf { y } } ( n ) \hat { \mathbf { x } } ( n ) - \hat { \zeta } ( n ) + \mu \hat { \mathbf { w } } ( n ) ) 
	\\ 
	& - \frac { \hat { \mathbf { x }_0 } ( n ) } { \mu b } \left( \hat{s}_{\mathbf{x}}(n) \hat{y}(n) - \hat { s } _ { \zeta } ( n ) + \mu \hat { s } _ { \mathbf { w } } ( n ) \right) 
	\end{aligned} \ , 
	\end{equation}
\noindent where $\hat{s}_{\mathbf{x}}(n) = \hat { \mathbf { x } } ( n ) ^{\top} \hat { \mathbf { x } } ( n )$, $\hat { s } _ { \zeta } ( n ) = \hat { \mathbf { x } } ( n ) ^{\top} \mathbf{\hat{\zeta}}$, $\hat { s } _ { \mathbf { w } } ( n ) = \hat { \mathbf { x } } ( n ) ^{\top} \mathbf{\hat{w}}$ and $b = \hat { \mathbf { x } } ( n ) ^{\top} \hat { \mathbf { x } } ( n ) + \mu$.

Lagrangian parameter is updated as follows:
	\begin{equation}
	\mathbf{\hat{\zeta}} _ {j+1} = \mathbf{\hat{\zeta}} _ {j} + \mu \left(\mathbf{\hat{g}} ^ {*} _ {j+1} - \mathbf{\hat{w}} ^ {*} _ {j+1} \right) \ , 
	\end{equation}
\noindent where subscript $j$ denotes the initial value or the value in the last iteration and subscript $j+1$ denotes the value in current iteration. So practically, we use the last(or initial) value of $\mathbf{\zeta}$ and the current solution to the aforementioned subproblems $\mathbf{\hat{g}}^*$ and $\mathbf{\hat{w}}^*$ to update the Lagrangian parameter. Note that $\mathbf{\hat{w}}^{*}_{j+1} = \left(\mathbf{I}_D \otimes \mathbf{FB}^{\top}\right) \mathbf{{w}}^{*}_{j+1}$.

\begin{center}
	\begin{figure}[!b]
		\centering
		\includegraphics[scale=0.3]{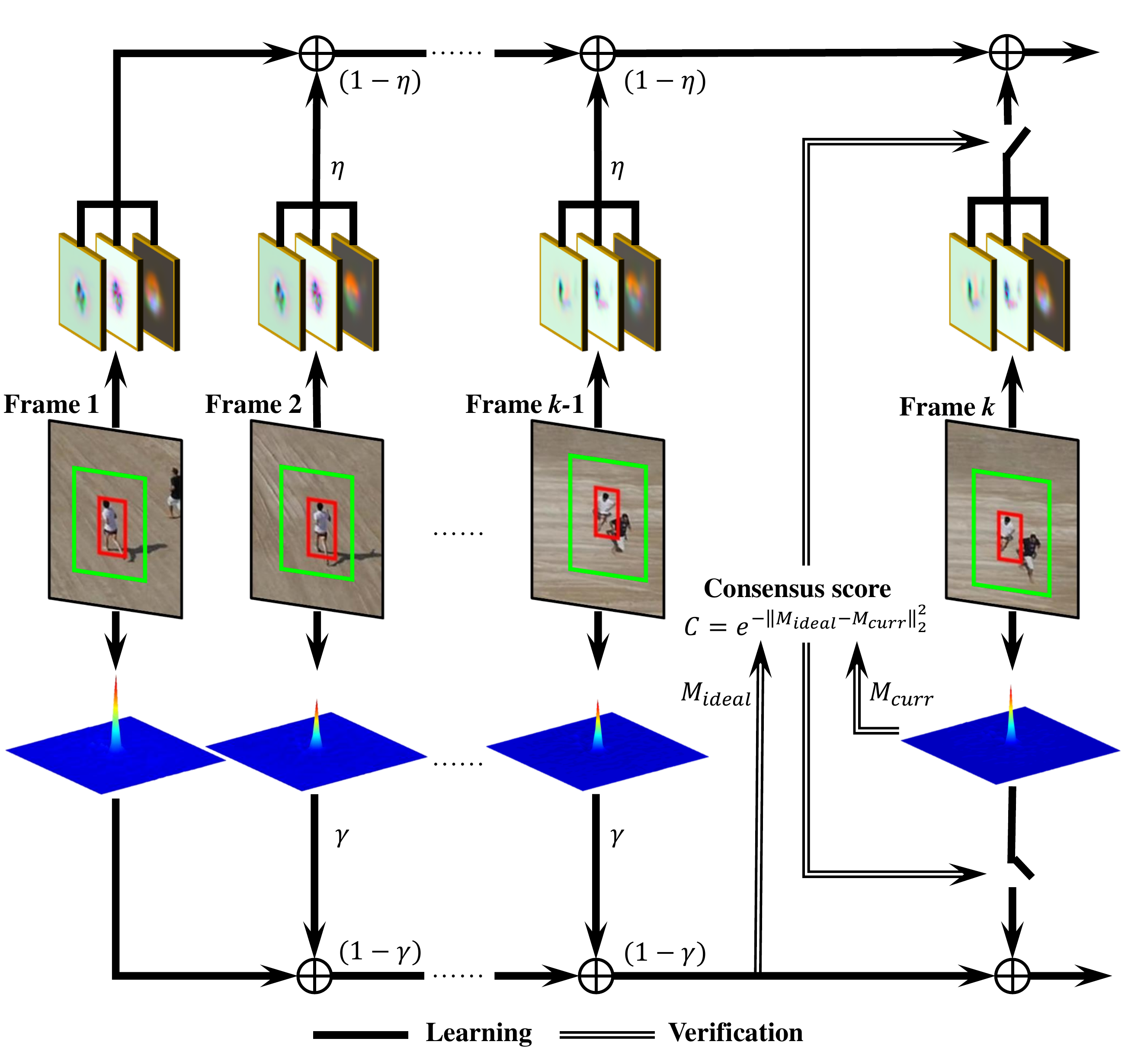}
		\caption{Schematic diagram of verification in model update process. Top path is for learning of object appearance model (features extracted from three layers of CNN). Bottom path is for learning of response map.}
		\label{img:verification}
	\end{figure}
\end{center}
\subsection{Model update}
\label{subsec:MODELUPDATE}
Most currently existing trackers update appearance models and CFs on a frame-by-frame basis, ignoring the possibility that tracking result at current frame can be inaccurate. Blindly updating appearance model and correlation filters would cause the tracker to learn wrongly detected objects. Since response map reveals similarity between object and model, it is used in our work to estimate consensus so that sudden changes will not be learned. 

In the first frame when UAV has detected an object, a new model is generated and an initial correlation filter is learned in this frame. The correlation filter learned is then directly used to convolute with the detected object to generate an ideal response map. In the other frames, the following criterion are used to verify the consensus score $\mathbf{C}$ of that frame:
	\begin{equation}
	\mathbf{C} = \mathbf{e} ^ {- \lVert \mathbf{M}_{\text{ideal}} - \mathbf{M}_{\text{curr}} \rVert _2 ^ 2} \ ,
	\end{equation}

\noindent where $\mathbf{M}_{ideal}$ is the ideal response map and $\mathbf{M}_{curr}$ is the response map generated in the current frame. \\
\noindent\textbf{Remark 4}: Verification process in each frame (except the first) generates a consensus score and control the learning of both object appearance model and response map model. The process can be seen in Fig. \ref{img:verification}.

When consensus is high enough, appearance model $\mathbf{\hat{x}}$ and ideal response map $\mathbf{M}_{ideal}$ are updated to improve its robustness to pose, scale and illumination changes with learning rates respectively being $\eta$ and $\gamma$ as follows:
	\begin{equation}\label{update_strategy}
	\begin{aligned}
	&\mathbf{\hat{x}}^{i} = (1-\eta)\mathbf{\hat{x}}^{i-1} + \eta\mathbf{\hat{x}}_{\text{curr}} \\
		&\mathbf{M}_{\text{ideal}}^{i} = (1-\gamma)\mathbf{M}_{\text{ideal}}^{i-1} + \gamma\mathbf{M}_{\text{curr}}
	\end{aligned} \ ,
	\end{equation}
	
\noindent where superscript $i$ denotes current frame and $i-1$ denotes the model learned in the last learning. Subscript $curr$ denotes the feature extracted or response map generated in the current frame. Note that we use $\mathbf{\hat{x}}$ updated here to compute parameters in section \ref{subsec:OPTIMIZATIONOPERATIONS}. 

\noindent\textbf{Remark 5}: Compared to other CF-based trackers, which only learns object appearance model, ideal response map is also learned and updated. In addition, when the estimated consensus score is considered high enough, a reinforcement technique is used to reinforce the learning of current model, that is, the learning rate in this frame will be temporally boosted. Details can be seen in Algorithm \ref{alg:BEVTtrackerflow}.
\subsection{Detection}
\label{subsec:DETECTION}
Object location in the frame is detected by applying filter updated in the last allowed learning phase (controlled by consensus score) to the newly extracted convolutional features on different resolutions of the searching area to estimate scale differences likewise to  \cite{danelljan2015ICCV}. Interpolation strategy is applied to maximize detection scores in each correlation result. The object's scale is estimated by the scale with maximum correlation score and location is then estimated according to this scale. 
\begin{center}
	\begin{figure}[!b]
		\centering
		\includegraphics[scale=0.33]{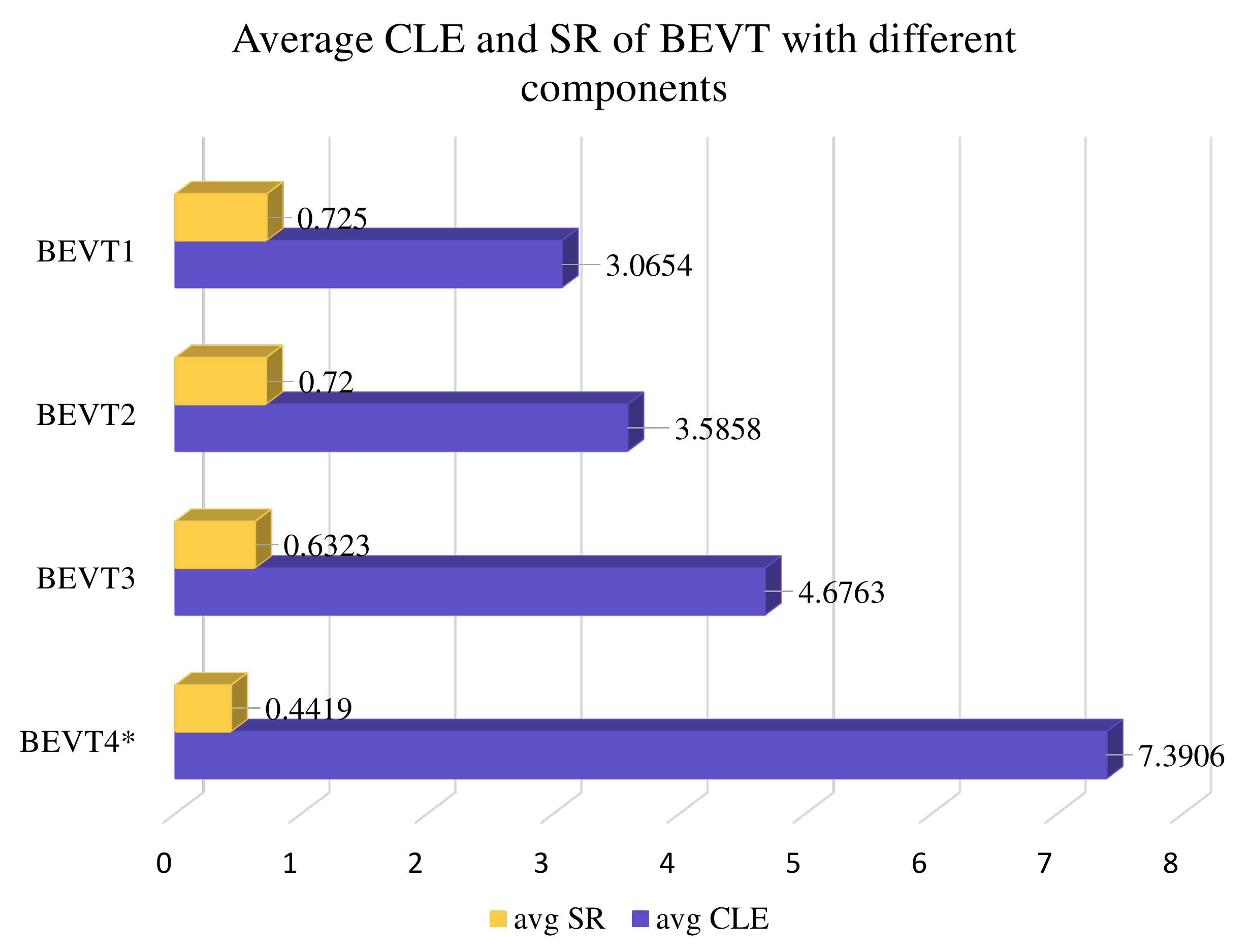}
		\caption{Comparison between BEVT with different modules on \textit{person2\_2} image sequence. BEVT1 is the full version, i.e. BEVT. BEVT2 is BEVT without response map verification. BEVT3 is BEVT2 without enhanced background learning. BEVT4 is BEVT3 using HOG feature instead of convolutional features. Note: because BEVT4 has lost the object from frame 180, so only the result of former 180 frames is shown.}
		\label{img:comp_components}
	\end{figure}
\end{center}
\subsection{Module evaluation}
In order to verify the effectiveness of the presented modules, the proposed tracker with different modules are tested on \textit{person2\_2} image sequence. 

Two evaluation criteria are employed to evaluate the performance of the proposed tracker, respectively center location error (CLE) and success rate (SR) based on one-pass evaluation (OPE).

CLE is defined as average Euclidean distance between the centers of detected bounding box of tracked objects and the manually annotated ground truth. SR shows the percentage of cases when the overlap area between annotated bounding boxes and detected ones are greater than a certain threshold. 

As is shown in Fig. \ref{img:comp_components}, every module added to the original framework would result in satisfying improvements in the  CLE and SR. 

In Section \ref{sec:EXPERIMENTS}, more exhaustive experiments were conducted to evaluate the performance of the proposed BEVT tracker.
\begin{algorithm}[!t]
	\caption{BEVT tracker}
	\label{alg:BEVTtrackerflow}
	\KwIn {\hspace{0.35cm}Object location on frame ${k-1}$, \\
		\hspace{1.45cm}Learned response model $\mathbf{M}_{\text{ideal}}^{k-1}$ }
	\KwOut {Estimated location on frame ${k}$}
	\For{$k=2$ to end}{
		Extract the search window in frame $i$ centered at object location on frame $k-1$\\
		Represent the extracted search window using convolutional features $\mathbf{\hat{x}}^{k}_{\text{detect}}$\\
		Convolute learned CF $\mathbf{\hat{g}}_{k-1}$ with $\mathbf{\hat{x}}^{k}_{\text{detect}}$ on different scales to generate $\mathbf{M}_{\text{curr}}$ and detect position\\
		\eIf{consensus score $C > \text{threshold}_{high}$}{$\eta$ = high\_learning\_rate}
		{\eIf{consensus score $C > \text{threshold}_{low}$}{$\eta$ = low\_learning\_rate}
			{Start detection of next frame $k+1$}}
		Extract convolutional features where $\mathbf{M}_{\text{curr}}$ is highest\\
		Update object and response map model by (\ref{update_strategy})\\
		Learn CF $\mathbf{\hat{g}}_{k}$ by (\ref{subproblem_w}), (\ref{subproblem_g_result}) and (\ref{Lagragian_format})
	}
\end{algorithm} 
\begin{figure*}[!ht]
	\centering
	\includegraphics[scale=0.55]{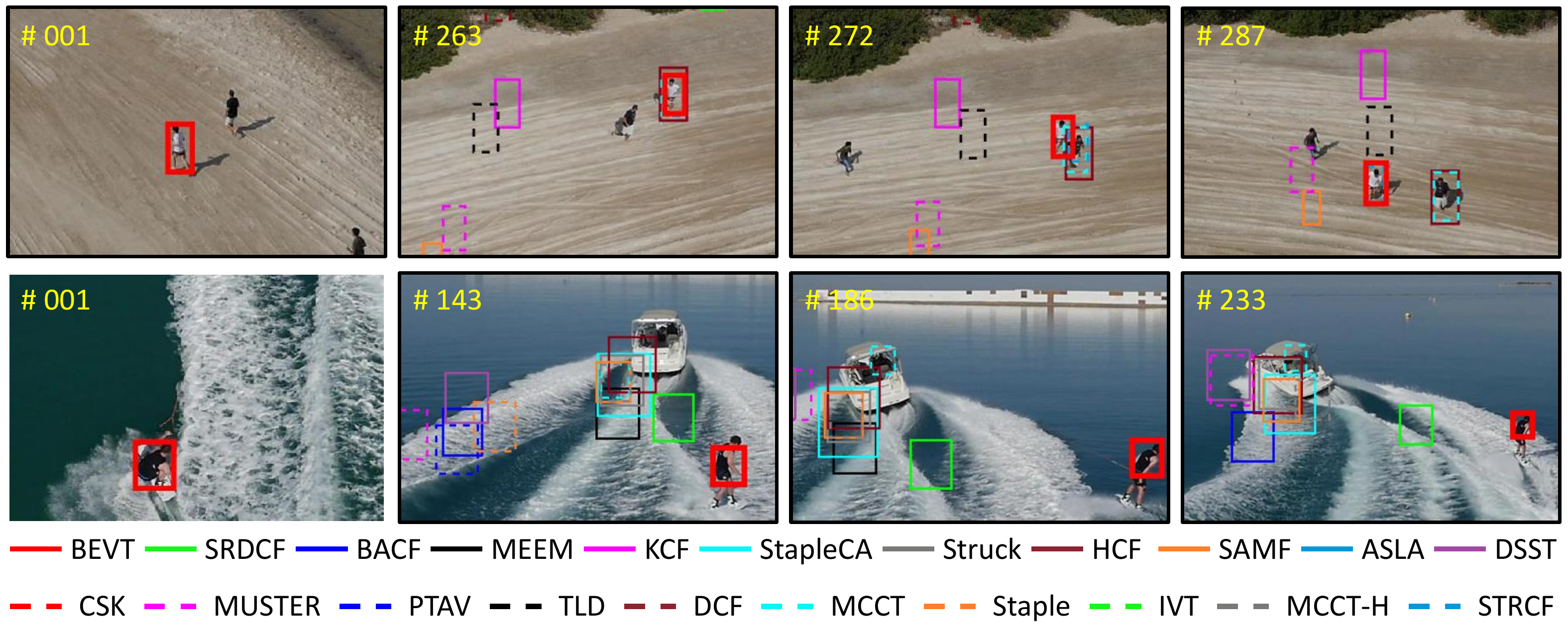}
	\caption{Examples of UAV tracking results. The first row is \textit{waterboard4} and the second is \textit{group2\_2} image sequences. Tracking code and video can be found here: \url{http://github.com/vision4robotics/BEVT-tracker} and \url{https://youtu.be/wjMQmx1_qkw}}
	\label{img:experiment_fig}
\end{figure*}
\section{EXPERIMENTS}
\label{sec:EXPERIMENTS}
The proposed BEVT tracker is evaluated extensively and thoroughly on 100 challenging UAV image sequences from well-known UAV123\_10fps \cite{mueller2016ECCV}, which is specifically designed and annotated for UAV tracking including exhaustive scenarios such as person, vehicle and boat following as well as UAV following. In this section, evaluation criteria is explained and comparisons are made between the state-of-the-art trackers. Some of the examples of UAV tracking results are provided in Fig. \ref{img:experiment_fig}.
\subsection{Evaluation criteria}
Besides aforementioned CLE and SR, two plots based on two criteria are employed to visualize the evaluation result. Precision plot (PP) is adopted to demonstrate percentage of frames whose CLE is within a certain threshold. Conventionally, threshold of 20 pixels is chosen to give the overall precision score \cite{mueller2016ECCV}. Success plot (SP) demonstrates the percentage of frames whose overlap score (OS) is within a certain threshold and area under curve (AUC) is adopted to give overall overlap score \cite{mueller2016ECCV}.\\
\noindent\textbf{Remark 6}: The experiments were carried out strictly according to standard procedure and standard evaluation criteria in the visual object tracking field.
\begin{figure}[!b]
	\centering
	\includegraphics[scale=0.31]{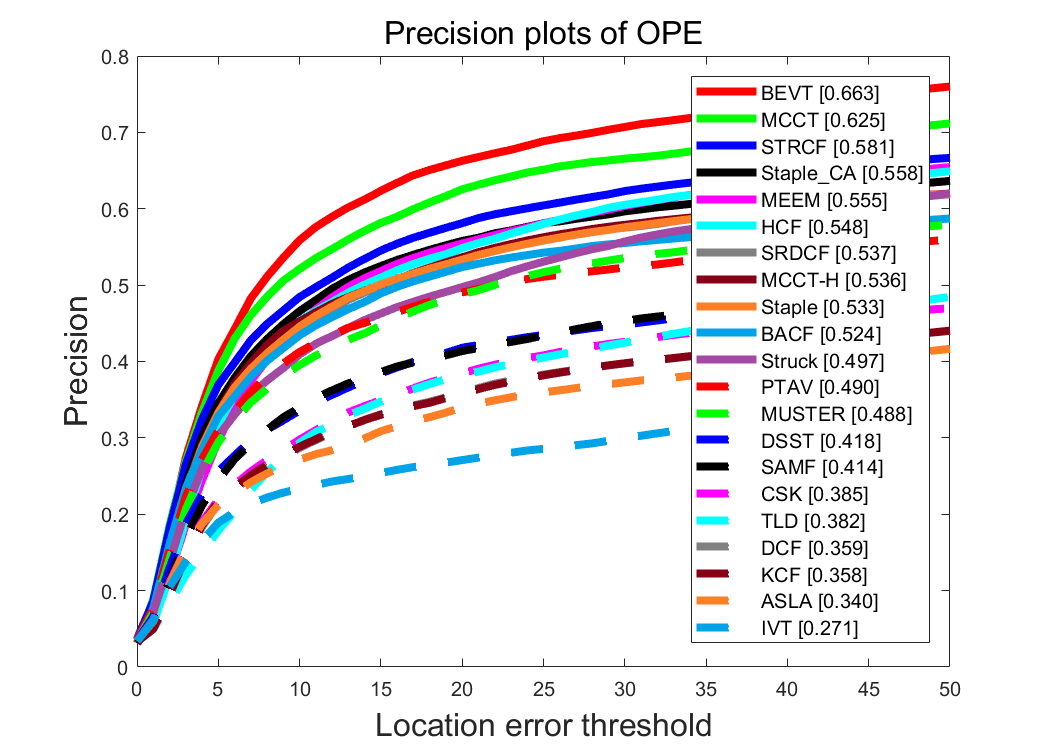}
	\caption{Precision plots of 21 trackers on 100 challenging UAV image sequences. The BEVT tracker has a superiority of respectively 3.8\% and 8.2\% in comparison with the second best tracker MCCT (2018 CVPR) and third best tracker STRCF (2018 CVPR).}
	\label{img:comp_sec_1_pp}
\end{figure}
\begin{figure}[!b]
	\centering
	\includegraphics[scale=0.31]{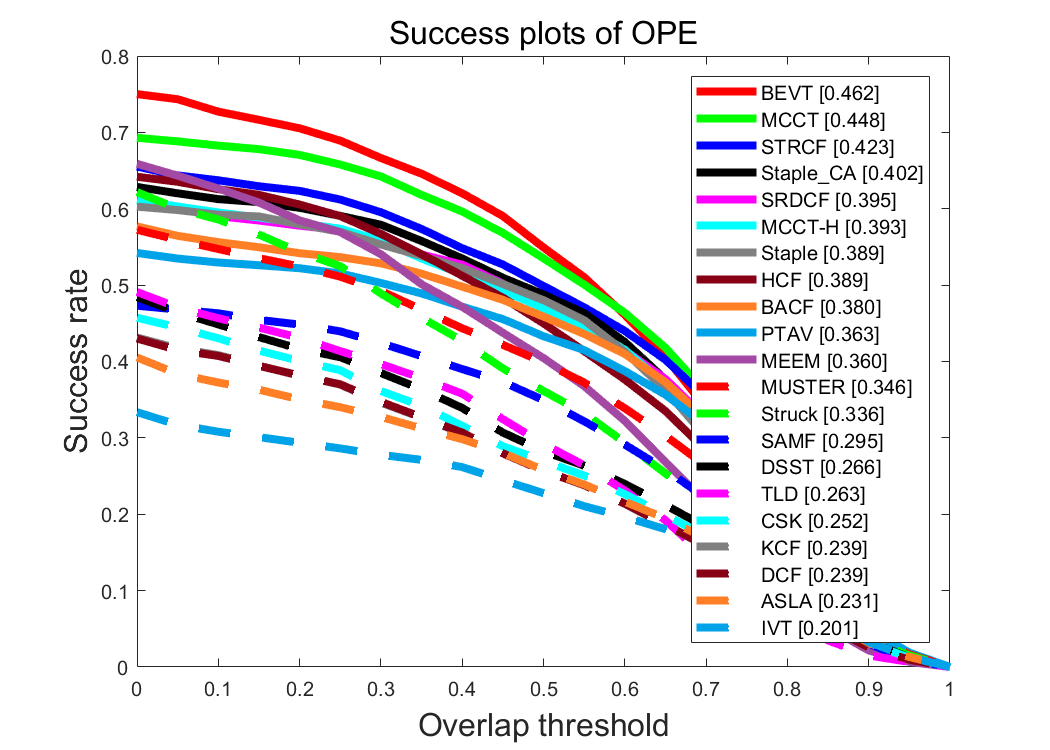}
	\caption{AUCs of 21 trackers on 100 challenging UAV image sequences. The BEVT tracker has a superiority of respectively 1.4\% and 5.0\% in comparison with the second best tracker MCCT (2018 CVPR) and third best tracker STRCF (2018 CVPR).}
	\label{img:comp_sec_1_auc}
\end{figure}

\subsection{Comparison with state-of-the-art trackers}
\begin{figure}[!t]
	\centering
	\includegraphics[scale=0.28]{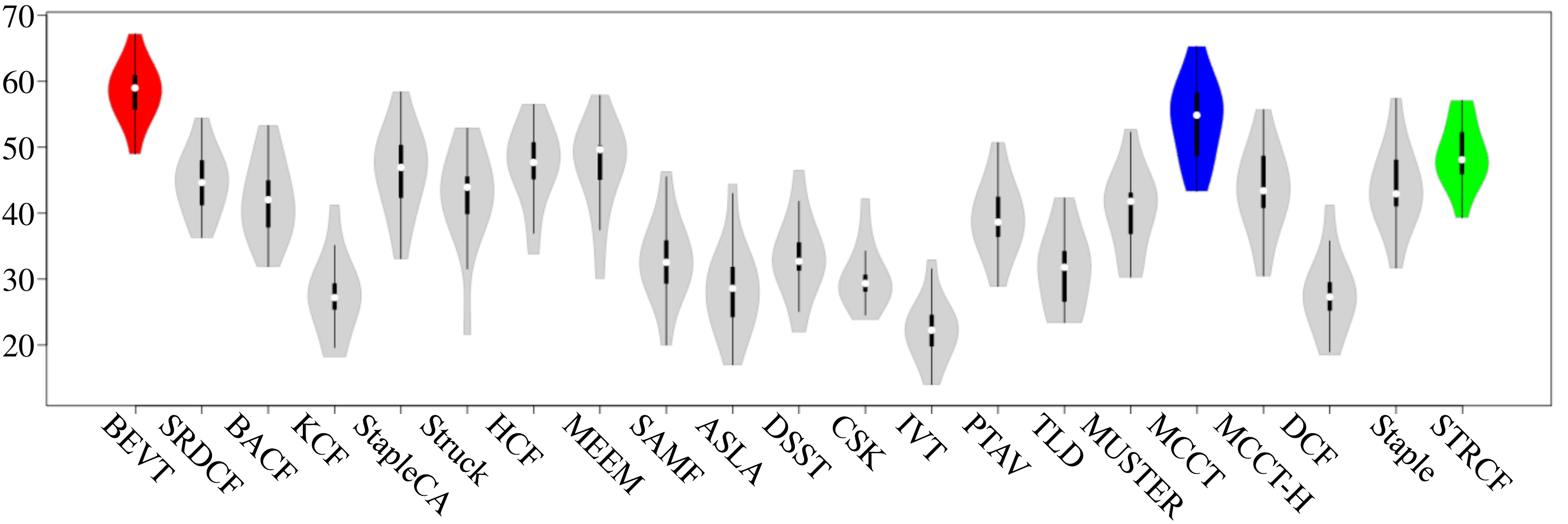}
	\caption{12 attributes based on PP results of 21 trackers. {\color{red}Red}, {\color{blue}blue} and {\color{green}green} fonts respectively indicates the best, second best and third best performances among all trackers.}
	\label{img:violin_plot_pp}
\end{figure}

\begin{figure}[!t]
	\centering
	\includegraphics[scale=0.28]{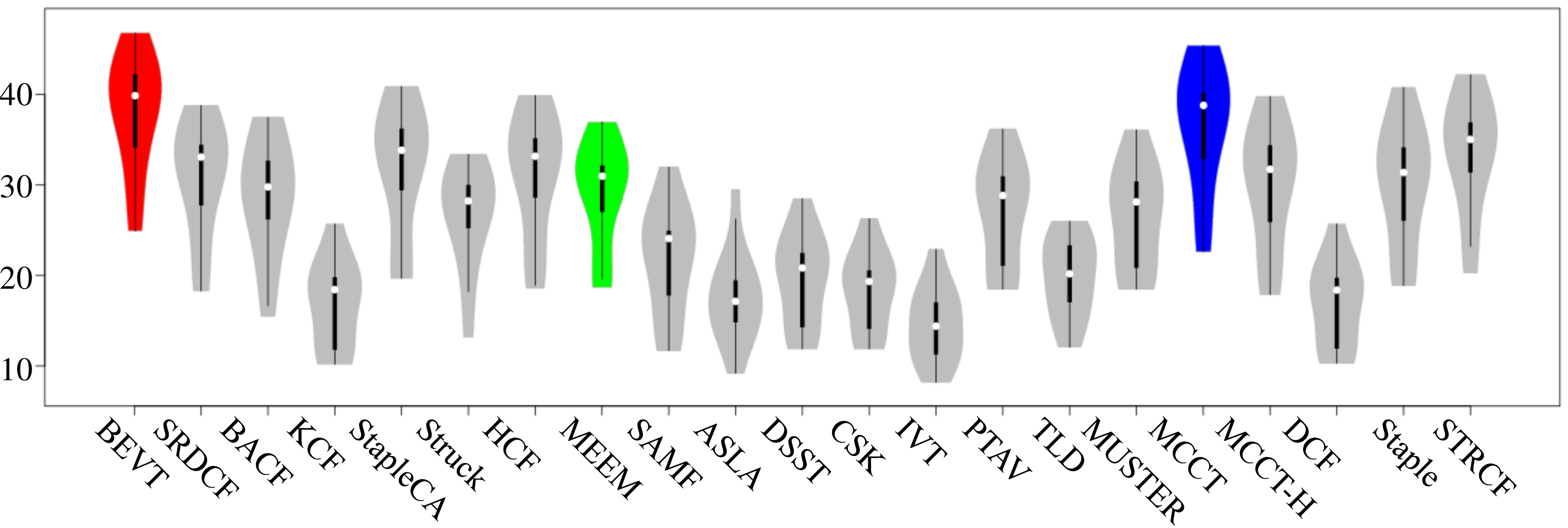}
	\caption{12 attributes based on AUC results of 21 trackers. {\color{red}Red}, {\color{blue}blue} and {\color{green}green} fonts respectively indicates the best, second best and third best performances among all trackers.}
	\label{img:violin_plot_auc}
\end{figure}

In this section, the tracking performances of the proposed tracker is demonstrated based on the aforementioned criteria. In oder to achieve a comprehensive evaluation, the BEVT tracker is horizontally compared with 20 state-of-the-art trackers, i.e. HCF \cite{ma2015ICCV}, BACF \cite{kiani2017ICCV}, KCF \cite{henriques2015TPAMI}, StapleCA \cite{mueller2017CVPR}, Staple \cite{bertinetto2016CVPR}, Struck \cite{hare2016PAMI}, ASLA \cite{jia2012CVPR}, CSK \cite{henriques2012ECCV}, IVT \cite{ross2008IJCV}, PTAV \cite{fan2017ICCV}, TLD \cite{kalal2012PAMI}, MUSTER \cite{hong2015CVPR}, MCCT \cite{wang2018CVPR}, MCCT-H \cite{wang2018CVPR}, DCF \cite{henriques2015TPAMI}, STRCF \cite{li2018CVPR}, SRDCF \cite{danelljan2015ICCV}, MEEM \cite{zhang2014ECCV}, SAMF \cite{li2014ECCV}, DSST \cite{danelljan2014BMVC}, on 100 challenging UAV image sequences captured from a low-altitude aerial perspective \cite{mueller2016ECCV}, which are nearly identical to the real-world UAV following cases. Fig. \ref{img:comp_sec_1_pp} and Fig. \ref{img:comp_sec_1_auc} shows all the results of PPs and AUCs of 21 trackers on 100 challenging UAV image sequences. In terms of precision and success ratio, the proposed BEVT tracker has demonstrated a superiority in overall performance. \\
\noindent\textbf{Remark 7}: All trackers are implemented with MATLAB
R2017a and all the experiments were run on the computer with
an i7-8700K processor (3.7GHz), 48GB RAM and NVIDIA
Quadro P2000 GPU.
\begin{center}
	\begin{table*}[!t]
		\scriptsize
		\centering
		\caption{Scores of PP (threshold at 20 pixels). {\color{red}Red} and {\color{blue}blue} fonts respectively indicates BEVT's best and second best performances among all trackers.}
		\begin{tabular}{l|c|c|c|c|c|c|c|c|c|c|c|c}
			&\textbf{ARC}&\textbf{BC}&\textbf{CM}&\textbf{FM}&\textbf{FOC}&\textbf{IV}&\textbf{LR}&\textbf{OV}&\textbf{POC}&\textbf{SV}&\textbf{SOB}&\textbf{VC}\\\hline
			\textbf{BEVT } & \textcolor[rgb]{ 1,  0,  0}{59.5} & \textcolor[rgb]{ 1,  0,  0}{54.4} & \textcolor[rgb]{ 1,  0,  0}{67.3} & \textcolor[rgb]{ 1,  0,  0}{56.7} & \textcolor[rgb]{ 1,  0,  0}{49.1} & \textcolor[rgb]{ 1,  0,  0}{58.7} & \textcolor[rgb]{ 1,  0,  0}{55.2} & \textcolor[rgb]{ 1,  0,  0}{56.1} & \textcolor[rgb]{ 1,  0,  0}{59.9} & \textcolor[rgb]{ 1,  0,  0}{62.5} & \textcolor[rgb]{ 0,  0,  1}{62.4} & \textcolor[rgb]{ 1,  0,  0}{60.5} \\
			\textbf{Best of other trackers } & 55.3 & 51.3 & 60.7 & 48.6 & 45.3 & 55.7 & 46.7 & 54 & 58.2 & 58.4 & 65.3 & 54.4 \\
			\textbf{Avg. of other trackers } & 39.565 & 34.68 & 41.33 & 29.465 & 34.915 & 37.035 & 36.33 & 36.515 & 42.15 & 43.43 & 50.45 & 38.555 \\\hline
		\end{tabular}%
		\label{tab:PP}%
	\end{table*}%
\end{center}
\begin{center}
	\begin{table*}[!t]
		\scriptsize
		\centering
		\caption{Scores of SR (AUC). {\color{red}Red} and {\color{blue}blue} fonts respectively indicates BEVT's best and second best performances among all trackers.}
		\begin{tabular}{l|c|c|c|c|c|c|c|c|c|c|c|c}
			\hline
			&\textbf{ARC}&\textbf{BC}&\textbf{CM}&\textbf{FM}&\textbf{FOC}&\textbf{IV}&\textbf{LR}&\textbf{OV}&\textbf{POC}&\textbf{SV}&\textbf{SOB}&\textbf{VC}\\\hline
			\textbf{BEVT } & \textcolor[rgb]{ 1,  0,  0}{39.5} & \textcolor[rgb]{ 1,  0,  0}{34.3} & \textcolor[rgb]{ 1,  0,  0}{46.9} & \textcolor[rgb]{ 1,  0,  0}{34.2} & \textcolor[rgb]{ 1,  0,  0}{25} & \textcolor[rgb]{ 1,  0,  0}{40.5} & \textcolor[rgb]{ 1,  0,  0}{28.4} & \textcolor[rgb]{ 1,  0,  0}{39.3} & \textcolor[rgb]{ 1,  0,  0}{40.4} & \textcolor[rgb]{ 1,  0,  0}{43} & \textcolor[rgb]{ 0,  0,  1}{44.1}  & \textcolor[rgb]{ 1,  0,  0}{42} \\
			\textbf{Best of other trackers } & 37.8 & 33.4 & 44.6 & 31.5 & 22.8 & 38.8 & 25.3 & 38.9 & 39.7 & 41.1 & 45.5 & 39.9 \\
			\textbf{Avg. of other trackers } & 26.875 & 22.235 & 29.775 & 18.275 & 17.025 & 26.055 & 17.94 & 26.33 & 28.255 & 29.97 & 34.16 & 28.06
			\\\hline
			
		\end{tabular}%
		\label{tab:SR}%
	\end{table*}%
\end{center}
Besides overall performance, the proposed tracker and 20 other top trackers also compare on 12 attributes categorized by 100 challenging UAV image sequences, i.e. aspect ratio change (ARC), background clutter (BC), camera motion (CM), fast motion (FM), full occlusion (FOC), illumination variation (IV), low resolution (LR), out-of-view (OV), partial occlusion (POC), scale variation (SV), similar object (SOB) and viewpoint change (VC) \cite{mueller2016ECCV}. Table \ref{tab:PP} and \ref{tab:SR} show the scores in CLE and AUC evaluation on mentioned 12 attributes. The proposed tracker has achieved the best performance in both CLE and AUC evaluation in all aspects but SOB, in which BEVT both achieved second place. Two violin plots Fig. \ref{img:violin_plot_pp} and Fig. \ref{img:violin_plot_auc} visualizes the results of PP and AUC of all evaluated trackers respectively. 

\noindent\textbf{Remark 8}: Among all the attributes that our tracker performs satisfactory, in the evaluation attribute CM, FM, LR and VC, the proposed BEVT tracker outperforms the other tracker substantially. The superiority of performance of BEVT tracker in CM, FM and VC probably results from the large search area combined with strong background noise suppressing ability since all these scenarios include large-scale motion of tracked objects. Due to the ability of consensus verification module to verify tracking result and interfere with learning process, BEVT tracker is able to learn a more robust and accurate object appearance model even when there are few information and many noises in LR scenarios. All these scenarios appear extremely frequently. Our lead in FM, CM and VC demonstrates the trackers ability to follow the object in higher relative speed and the lead in LR means the proposed tracker can follow smaller objects (farther) or in bad weather. It can be thus concluded that our tracker are more suitable than other trackers for UAV to perform object tracking at higher speed and in harsher environments.

\subsection{Limitations}
\textbf{Background clutter performance:} Due to the inclusion of more real background information, similar objects are more likely to be included in the detection phase. Therefore, the BEVT tracker's performance is limited when there is similar objects in the background, especially in cases when the target is occluded, as is shown in Table \ref{tab:PP} and Table \ref{tab:SR}. 

\textbf{Spatial penalization disturbance: }What's more, in order to simplify calculation, matrix multiplication was used instead of element-wise multiplication of penalization term $\mathbf{p}$ and correlation filter $\mathbf{w}$ to enforce spatial penalization. This would introduce unwanted disturbance in spatial punishment. Therefore, a bowl-shaped penalization term might not be the best penalization term. A more suitable penalization matrix can be more carefully selected in the future.

\textbf{Speed: }In addition, the BEVT tracker is implemented on MATLAB platform with no optimization. Therefore, the frame per second achieves only on average 0.653 on the 100 challenging UAV image sequences on the platform mentioned before. Fortunately, the used DJI S1000+ is capable of carrying high-performance GPU and CPU. With proper optimization for GPU and implementation of parallel computation, BEVT tracker can be applied to UAV tracking.

\section{CONCLUSIONS}
\label{sec:CONCLUSIONS}

In this work, tracker with online enhanced background learning and multi-frame consensus verification (response map verification) is proposed. Enhanced background learning solves the boundary effect of BEVT tracker and reduces the noise introduced by background. Convolutional features improve the description of object by making it more comprehensive with spatial details and semantic information. The introduced consensus verification further adds to the robustness of learned model. The BEVT tracker uses the ADMM to optimize the calculation of correlation filter. After performing exhaustive experiments and comparisons on 100 challenging UAV image sequences, it is proven that despite our limitation in speed, the best tracking performance is achieved among all state-of-the-art trackers. Addressing boundary effect and strengthening representation of models enables UAV to track more complex objects in complicated environments. In our perspective, the results of this work further improves the CF framework in terms of limitation in boundary effect and extend the implementation of response map, thus promoting object tracking for UAV scenarios.
\section*{APPENDIX}
\label{APPENDIX}

Here is the deduction from (\ref{matrix_form_of_objective}) to (\ref{simplification_of_objective}). The summation term in the first norm in (\ref{matrix_form_of_objective}) can be simplified as follows:

	\begin{equation}
	\begin{aligned}
	&\sum_{d=1}^D \mathbf{w} ^{d\top} \mathbf{Bx} ^d[\Delta\tau_n] \\
	& =\left[\begin{matrix}\mathbf{w} ^{1\top}\mathbf{B}&\mathbf{w} ^{2\top}\mathbf{B}&\cdots &\mathbf{w} ^{D\top}
	\mathbf{B}\end{matrix}\right] \cdot \\
	&\quad \quad \left[ \begin{matrix} \mathbf{x} ^1[\Delta\tau_n]^{\top}\  \mathbf{x} ^2[\Delta\tau_n]^{\top}\  \cdots \  \mathbf{x} ^D[\Delta\tau_n]^{\top}\ \end{matrix}\right]^{\top}
	\\
	& =\left[\begin{matrix}\mathbf{w} ^{1\top}&\mathbf{w} ^{2\top}&\cdots &\mathbf{w} ^{D\top}\end{matrix}\right] \left( \mathbf{I}_D \otimes \mathbf{B}\right) \cdot \\
	&\quad \quad \left[ \begin{matrix} \mathbf{x} ^1[\Delta\tau_n]^{\top}\  \mathbf{x} ^2[\Delta\tau_n]^{\top}\ \cdots \  \mathbf{x} ^D[\Delta\tau_n]^{\top}\ \end{matrix}\right]^{\top}
	\end{aligned} \ ,
	\end{equation}

\noindent which is a scalar, so it can be further simplified as
	\begin{equation}
	\begin{aligned}
	& \sum_{k=1}^K \mathbf{h} _k^T \mathbf{Px} _k[\Delta\tau_j]\\
	& = \mathbf{x}[\Delta\tau_n] \left( \mathbf{I}_D \otimes \mathbf{B}\right) \mathbf{h} \\ 
	\end{aligned} \ ,	
	\end{equation}

\noindent where $\mathbf{x}[\Delta\tau_n] = \left[ \begin{matrix} \mathbf{x} ^1[\Delta\tau_n]^{\top}& \mathbf{x} ^2[\Delta\tau_n]^{\top} & \cdots & \mathbf{x} _D[\Delta\tau_n]^{\top}\end{matrix}\right]$ and $\mathbf{h} = \left[\begin{matrix}\mathbf{w} ^{1\top}  \mathbf{w} ^{2\top}  \cdots  \mathbf{w} ^{D\top}\end{matrix}\right] ^{\top}$.

Therefore, (\ref{matrix_form_of_objective}) can be simplified as follows: 
	\begin{equation}
	\mathcal { E } ( \mathbf { w } ) = \frac{1}{2} \sum_{n=1}^N \lVert \mathbf{y}(n) - 
	\mathbf{x}[\Delta\tau_n]  \left(\mathbf{I}_D \otimes \mathbf{B}^{\top}\right) \mathbf{w}\rVert^2 + 
	\lVert \mathbf{\tilde{p}w} \rVert ^2 \ ,
	\end{equation}

\noindent which is (\ref{simplification_of_objective}) after definition in Section \ref{sec:PROPOSEDTRACKINGAPPROACH}.

\section*{ACKNOWLEDGMENT}

This work is supported by the National Natural Science Foundation of China (No. 61806148) and the Fundamental Research Funds for the Central Universities (No. 22120180009).

\bibliographystyle{IEEEtran}
\bibliography{IROS2019}

\begin{thebibliography}{10}
\providecommand{\url}[1]{#1}
\csname url@rmstyle\endcsname
\providecommand{\newblock}{\relax}
\providecommand{\bibinfo}[2]{#2}
\providecommand\BIBentrySTDinterwordspacing{\spaceskip=0pt\relax}
\providecommand\BIBentryALTinterwordstretchfactor{4}
\providecommand\BIBentryALTinterwordspacing{\spaceskip=\fontdimen2\font plus
\BIBentryALTinterwordstretchfactor\fontdimen3\font minus
  \fontdimen4\font\relax}
\providecommand\BIBforeignlanguage[2]{{%
\expandafter\ifx\csname l@#1\endcsname\relax
\typeout{** WARNING: IEEEtran.bst: No hyphenation pattern has been}%
\typeout{** loaded for the language `#1'. Using the pattern for}%
\typeout{** the default language instead.}%
\else
\language=\csname l@#1\endcsname
\fi
#2}}

\bibitem{fu2014ICRA}
C.~Fu, A.~Carrio, M.~A. Olivares-Mendez, R.~Suarez-Fernandez, and P.~Campoy,
  ``Robust real-time vision-based aircraft tracking from unmanned aerial
  vehicles,'' in \emph{2014 ieee international conference on robotics and
  automation (ICRA)}, 2014, pp. 5441--5446.

\bibitem{olivares2015Sensors}
M.~Olivares-Mendez, C.~Fu, P.~Ludivig, T.~Bissyand{\'e}, S.~Kannan, M.~Zurad,
  A.~Annaiyan, H.~Voos, and P.~Campoy, ``Towards an autonomous vision-based
  unmanned aerial system against wildlife poachers,'' \emph{Sensors}, vol.~15,
  no.~12, pp. 31\,362--31\,391, 2015.

\bibitem{mueller2016IROS}
M.~Mueller, G.~Sharma, N.~Smith, and B.~Ghanem, ``Persistent aerial tracking
  system for uavs,'' in \emph{2016 IEEE/RSJ International Conference on
  Intelligent Robots and Systems (IROS)}.\hskip 1em plus 0.5em minus
  0.4em\relax IEEE, 2016, pp. 1562--1569.

\bibitem{mueller2016ECCV}
M.~Mueller, N.~Smith, and B.~Ghanem, ``A benchmark and simulator for uav
  tracking,'' in \emph{European conference on computer vision}.\hskip 1em plus
  0.5em minus 0.4em\relax Springer, 2016, pp. 445--461.

\bibitem{kiani2017ICCV}
H.~Kiani~Galoogahi, A.~Fagg, and S.~Lucey, ``Learning background-aware
  correlation filters for visual tracking,'' in \emph{Proceedings of the IEEE
  International Conference on Computer Vision}, 2017, pp. 1135--1143.

\bibitem{danelljan2015ICCV}
M.~Danelljan, G.~Hager, F.~Shahbaz~Khan, and M.~Felsberg, ``Learning spatially
  regularized correlation filters for visual tracking,'' in \emph{Proceedings
  of the IEEE international conference on computer vision}, 2015, pp.
  4310--4318.

\bibitem{henriques2015TPAMI}
J.~F. Henriques, R.~Caseiro, P.~Martins, and J.~Batista, ``High-speed tracking
  with kernelized correlation filters,'' \emph{IEEE transactions on pattern
  analysis and machine intelligence}, vol.~37, no.~3, pp. 583--596, 2015.

\bibitem{qi2016CVPR}
Y.~Qi, S.~Zhang, L.~Qin, H.~Yao, Q.~Huang, J.~Lim, and M.-H. Yang, ``Hedged
  deep tracking,'' in \emph{Proceedings of the IEEE conference on computer
  vision and pattern recognition}, 2016, pp. 4303--4311.

\bibitem{ma2015ICCV}
C.~Ma, J.-B. Huang, X.~Yang, and M.-H. Yang, ``Hierarchical convolutional
  features for visual tracking,'' in \emph{Proceedings of the IEEE
  international conference on computer vision}, 2015, pp. 3074--3082.

\bibitem{wang2015ICCV}
L.~Wang, W.~Ouyang, X.~Wang, and H.~Lu, ``Visual tracking with fully
  convolutional networks,'' in \emph{Proceedings of the IEEE international
  conference on computer vision}, 2015, pp. 3119--3127.

\bibitem{choi2017CVPR}
J.~Choi, H.~Jin~Chang, S.~Yun, T.~Fischer, Y.~Demiris, and J.~Young~Choi,
  ``Attentional correlation filter network for adaptive visual tracking,'' in
  \emph{Proceedings of the IEEE conference on computer vision and pattern
  recognition}, 2017, pp. 4807--4816.

\bibitem{wang2017CVPR}
M.~Wang, Y.~Liu, and Z.~Huang, ``Large margin object tracking with circulant
  feature maps,'' in \emph{Proceedings of the IEEE Conference on Computer
  Vision and Pattern Recognition}, 2017, pp. 4021--4029.

\bibitem{van2009TIP}
J.~Van De~Weijer, C.~Schmid, J.~Verbeek, and D.~Larlus, ``Learning color names
  for real-world applications,'' \emph{IEEE Transactions on Image Processing},
  vol.~18, no.~7, pp. 1512--1523, 2009.

\bibitem{danelljan2014BMVC}
M.~Danelljan, G.~H{\"a}ger, F.~Khan, and M.~Felsberg, ``Accurate scale
  estimation for robust visual tracking,'' in \emph{British Machine Vision
  Conference, Nottingham, September 1-5, 2014}.\hskip 1em plus 0.5em minus
  0.4em\relax BMVA Press, 2014.

\bibitem{simonyan2014arXiv}
K.~Simonyan and A.~Zisserman, ``Very deep convolutional networks for
  large-scale image recognition,'' \emph{arXiv preprint arXiv:1409.1556}, 2014.

\bibitem{boyd2011FTML}
S.~Boyd, N.~Parikh, E.~Chu, B.~Peleato, J.~Eckstein, \emph{et~al.},
  ``Distributed optimization and statistical learning via the alternating
  direction method of multipliers,'' \emph{Foundations and
  Trends{\textregistered} in Machine learning}, vol.~3, no.~1, pp. 1--122,
  2011.

\bibitem{sherman1950AMS}
J.~Sherman and W.~J. Morrison, ``Adjustment of an inverse matrix corresponding
  to a change in one element of a given matrix,'' \emph{The Annals of
  Mathematical Statistics}, vol.~21, no.~1, pp. 124--127, 1950.

\bibitem{mueller2017CVPR}
M.~{Mueller}, N.~{Smith}, and B.~{Ghanem}, ``Context-aware correlation filter
  tracking,'' in \emph{2017 IEEE Conference on Computer Vision and Pattern
  Recognition (CVPR)}, July 2017, pp. 1387--1395.

\bibitem{bertinetto2016CVPR}
L.~Bertinetto, J.~Valmadre, S.~Golodetz, O.~Miksik, and P.~H. Torr, ``Staple:
  Complementary learners for real-time tracking,'' in \emph{Proceedings of the
  IEEE conference on computer vision and pattern recognition}, 2016, pp.
  1401--1409.

\bibitem{hare2016PAMI}
S.~Hare, S.~Golodetz, A.~Saffari, V.~Vineet, M.-M. Cheng, S.~L. Hicks, and
  P.~H. Torr, ``Struck: Structured output tracking with kernels,'' \emph{IEEE
  transactions on pattern analysis and machine intelligence}, vol.~38, no.~10,
  pp. 2096--2109, 2016.

\bibitem{jia2012CVPR}
X.~Jia, H.~Lu, and M.-H. Yang, ``Visual tracking via adaptive structural local
  sparse appearance model,'' in \emph{2012 IEEE Conference on computer vision
  and pattern recognition}.\hskip 1em plus 0.5em minus 0.4em\relax IEEE, 2012,
  pp. 1822--1829.

\bibitem{henriques2012ECCV}
J.~F. Henriques, R.~Caseiro, P.~Martins, and J.~Batista, ``Exploiting the
  circulant structure of tracking-by-detection with kernels,'' in
  \emph{European conference on computer vision}.\hskip 1em plus 0.5em minus
  0.4em\relax Springer, 2012, pp. 702--715.

\bibitem{ross2008IJCV}
D.~A. Ross, J.~Lim, R.-S. Lin, and M.-H. Yang, ``Incremental learning for
  robust visual tracking,'' \emph{International journal of computer vision},
  vol.~77, no. 1-3, pp. 125--141, 2008.

\bibitem{fan2017ICCV}
H.~Fan and H.~Ling, ``Parallel tracking and verifying: A framework for
  real-time and high accuracy visual tracking,'' in \emph{Proceedings of the
  IEEE International Conference on Computer Vision}, 2017, pp. 5486--5494.

\bibitem{kalal2012PAMI}
Z.~Kalal, K.~Mikolajczyk, and J.~Matas, ``Tracking-learning-detection,''
  \emph{IEEE transactions on pattern analysis and machine intelligence},
  vol.~34, no.~7, pp. 1409--1422, 2012.

\bibitem{hong2015CVPR}
Z.~Hong, Z.~Chen, C.~Wang, X.~Mei, D.~Prokhorov, and D.~Tao, ``Multi-store
  tracker (muster): A cognitive psychology inspired approach to object
  tracking,'' in \emph{Proceedings of the IEEE Conference on Computer Vision
  and Pattern Recognition}, 2015, pp. 749--758.

\bibitem{wang2018CVPR}
N.~Wang, W.~Zhou, Q.~Tian, R.~Hong, M.~Wang, and H.~Li, ``Multi-cue correlation
  filters for robust visual tracking,'' in \emph{Proceedings of the IEEE
  Conference on Computer Vision and Pattern Recognition}, 2018, pp. 4844--4853.

\bibitem{li2018CVPR}
F.~Li, C.~Tian, W.~Zuo, L.~Zhang, and M.-H. Yang, ``Learning spatial-temporal
  regularized correlation filters for visual tracking,'' in \emph{Proceedings
  of the IEEE Conference on Computer Vision and Pattern Recognition}, 2018, pp.
  4904--4913.

\bibitem{zhang2014ECCV}
J.~Zhang, S.~Ma, and S.~Sclaroff, ``Meem: robust tracking via multiple experts
  using entropy minimization,'' in \emph{European conference on computer
  vision}.\hskip 1em plus 0.5em minus 0.4em\relax Springer, 2014, pp. 188--203.

\bibitem{li2014ECCV}
Y.~Li and J.~Zhu, ``A scale adaptive kernel correlation filter tracker with
  feature integration,'' in \emph{European conference on computer
  vision}.\hskip 1em plus 0.5em minus 0.4em\relax Springer, 2014, pp. 254--265.

\end{thebibliography}

\addtolength{\textheight}{-12cm}

\end{document}